\documentclass{article}

\usepackage{arxiv}

\usepackage{flushend}
\usepackage{lipsum}
\usepackage{amsmath,graphicx}
\usepackage[ruled,vlined]{algorithm2e}
\usepackage{enumitem}
\usepackage{booktabs, arydshln}
\usepackage{colortbl}
\usepackage{tabularx}
\usepackage{ragged2e}
\usepackage{array}
\usepackage{threeparttablex}
\usepackage{makecell}
\usepackage{overpic}
\usepackage{tikz}
\usepackage{color}
\usepackage[colorlinks]{hyperref}
\usepackage[nameinlink]{cleveref}
\definecolor{darkred}{rgb}{0.5,0,0}
\definecolor{darkgreen}{rgb}{0,0.5,0}
\definecolor{darkblue}{rgb}{0,0,0.5}
\definecolor{gray}{rgb}{0.35,0.35,0.35}
\hypersetup{ colorlinks,
linkcolor=darkblue,
filecolor=darkblue,
urlcolor=darkblue,
citecolor=black }
\usepackage{url}
\usepackage{amsfonts}       
\usepackage{nicefrac}       
\usepackage{microtype}      
\usepackage{bm}
\usepackage[scr=boondoxo]{mathalpha}
\usepackage{mathtools}

\usepackage{adjustbox}
\usepackage{setspace}
\usepackage{xargs}

\usepackage{caption}
\usepackage{subcaption}

\usepackage[utf8]{inputenc} 
\usepackage[T1]{fontenc}    
\usepackage{url}            
\usepackage{booktabs}       
\usepackage{amsfonts}       
\usepackage{nicefrac}       
\usepackage{microtype}      
\usepackage{graphicx}
\usepackage{natbib}
\usepackage{doi}

\renewcommand{\undertitle}{}
\newcommand{\ccclr}{C$^3$LR}

\newcolumntype{Y}{>{\centering\arraybackslash}X}
\makeatletter
\def\adl@drawiv#1#2#3{%
        \hskip.5\tabcolsep
        \xleaders#3{#2.5\@tempdimb #1{1}#2.5\@tempdimb}%
                #2\z@ plus1fil minus1fil\relax
        \hskip.5\tabcolsep}
\newcommand{\cdashlinelr}[1]{%
  \noalign{\vskip\aboverulesep
           \global\let\@dashdrawstore\adl@draw
           \global\let\adl@draw\adl@drawiv}
  \cdashline{#1}
  \noalign{\global\let\adl@draw\@dashdrawstore
           \vskip\belowrulesep}}
\makeatother

\title{SELF-SUPERVISED CLASS-COGNIZANT FEW-SHOT CLASSIFICATION}
\renewcommand\footnotemark{}

\author{\href{https://orcid.org/0000-0002-5123-5809}{\includegraphics[scale=0.06]{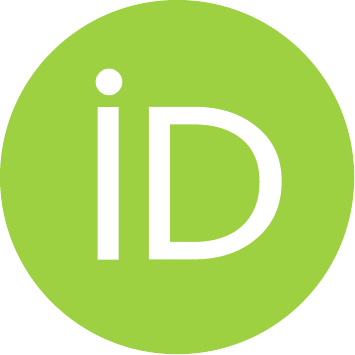}\hspace{1mm}Ojas Kishore Shirekar}$^{\dagger}$ \quad~\href{https://orcid.org/0000-0003-2254-6963 }{\includegraphics[scale=0.06]{orcid.pdf}\hspace{1mm}Hadi Jamali-Rad}$^{\dagger\star}$\thanks{The authors thank Delft University of Technology and Shell Global Solutions B.V. for permission to publish this work.}\\
{\textsuperscript{$\dagger$}Delft University of Technology (TU Delft), Delft, The Netherlands} \\{\textsuperscript{$\star$}Shell Global Solutions International B.V., Amsterdam, The Netherlands}\\ \small{\texttt{o.k.shirekar@student.tudelft.nl}},~
\small{\texttt{hadi.jamali-rad@shell.com}},\\
\small{\texttt{h.jamalirad@tudelft.nl}}}

\hypersetup{
pdftitle={SELF-SUPERVISED CLASS-COGNIZANT FEW-SHOT CLASSIFICATIO},
pdfsubject={cs.AI, cs.CV},
pdfauthor={ojas Kishore Shirekar, Hadi Jamali-Rad}
}

\begin{document}
\maketitle
\begin{abstract}
Unsupervised learning is argued to be the dark matter of human intelligence\footnote{\href{https://ai.facebook.com/blog/self-supervised-learning-the-dark-matter-of-intelligence/}{Yann LeCun's note; Meta AI blog post on self-supervised learning.}}. To build in this direction, this paper focuses on unsupervised learning from an abundance of unlabeled data followed by few-shot fine-tuning on a downstream classification task. To this aim, we extend a recent study on adopting contrastive learning for self-supervised pre-training by incorporating class-level cognizance through iterative clustering and re-ranking and by expanding the contrastive optimization loss to account for it. To our knowledge, our experimentation both in standard and cross-domain scenarios demonstrate that we set a new state-of-the-art (SoTA) in ($5$-way, $1$ and $5$-shot) settings of standard mini-ImageNet benchmark as well as the ($5$-way, $5$ and $20$-shot) settings of cross-domain CDFSL benchmark. Our code and experimentation can be found in our GitHub repository:\, \href{https://github.com/ojss/c3lr}{https://github.com/ojss/c3lr}\footnote{© 2022 IEEE. Personal use of this material is permitted. Permission from IEEE must be obtained for all other uses, in any current or future media, including reprinting/republishing this material for advertising or promotional purposes, creating new collective works, for resale or redistribution to servers or lists, or reuse of any copyrighted component of this work in other works.}. 

\end{abstract}
%
%
\section{Introduction}
\label{sec:intro}

Few-shot learning has received an upsurge of attention recently because it highlights a fundamental gap between human intelligence and data-hungry supervised deep learning methods. We humans can learn in a self-supervised fashion and/or with very little supervision. To tackle this challenge, few-shot classification is cast as the task of predicting class labels for a set of unlabeled data points (\textit{query set}) given only a small set of labeled ones (\textit{support set}). The query and support samples are typically drawn from the same distribution. Few-shot classification approaches are typically comprised of two sequential phases \citep{Medina2020Self-SupervisedClassification, chen2020simple, Ji2019UnsupervisedTraining, ye2020few}: (i) \emph{pre-training} on an abundant dataset (sometimes called ``base''), followed by (ii) \emph{fine-tuning} on an unseen dataset containing ``novel'' classes. Typically, the target classes in pre-training and fine-tuning phases are mutually exclusive. In this paper, we focus on self-supervised (also sometimes interchangeably called ``unsupervised'' in the literature) setting where we have no access to the class labels of the base dataset in the pre-training phase or their distribution. 

The art here is to devise a synthetic class label assignment technique and corresponding loss function in the pre-training phase to efficiently transfer the learning to the fine-tuning phase. To this aim, studies have proposed two different approaches. The first approach follows a \emph{meta-learning} strategy to create (synthetic) ``tasks'' similar to the the downstream episodic training in the fine-tuning phase \citep{Finn2017Model-agnosticNetworks, Hsu2018UnsupervisedMeta-Learning, Khodadadeh2018UnsupervisedClassification}. The second one follows some sort of \emph{transfer learning} approach, where a representation learning step in the pre-training phase is followed by episodic fine-tuning \citep{Medina2020Self-SupervisedClassification, goodemballneed2020, dhillon2019baseline}. In the latter case, typically a feature extractor (encoder) is trained using metric learning to capture the global structure of the unlabeled data. Next, a simple predictor (typically a linear layer) is adopted in conjunction with the extractor for quick adaptation to the novel classes in the fine-tuning phase. The better the feature extractor captures the global structure of the unlabeled data, the less the predictor requires training samples and the faster it adapts itself to the unseen classes in the fine-tuning phase. 

 Recent studies \citep{Medina2020Self-SupervisedClassification, chen2021self, dhillon2019baseline} demonstrate that the second approach based on transfer learning outperforms meta-learning based methods in cross-domain settings, where the training and novel classes come from totally different distributions. Their results also show that a properly-devised transfer learning based unsupervised approach comes pretty close to the performance of a fully supervised counterpart \citep{Medina2020Self-SupervisedClassification, Ji2019UnsupervisedTraining}, something that we will also confirm through experimentation. Most recently, a new state-of-the-art (SoTA) in self-supervised few shot classification has been set by extending the prototypical networks (ProtoNets) \citep{Snell2017PrototypicalLearning} using a \emph{contrastive} loss \citep{chen2020simple}. This approach (called ProtoTransfer \citep{Medina2020Self-SupervisedClassification}) constructs a contrastive metric embedding that clusters unlabeled prototypical samples and their augmentations. Inspired by this idea, we propose class-cognizant contrastive learning (C$^{3}$LR, \Cref{alg:pclr-c}) to further extend it to incorporate class-level insights from the global structure of data. This is done via an unsupervised iterative re-ranking and clustering step resulting in clusters of unlabeled embeddings followed by a modified contrastive loss now containing a term that specifically promotes this class-level global structure. Our experimentation demonstrates that C$^{3}$LR outperforms its predecessor ProtoTransfer in ($5$-way, $1$ and $5$-shot) settings of Ominglot \citep{Lake2015Human-levelInduction} and mini-Imagenet \citep{Vinyals2016MatchingLearning} benchmarks by about $1\%$ and $2\%+$, respectively. The performance improvement goes up to $4.5\%$ in the cross-domain setting of the CDFSL benchmark \citep{guo2019new}. As a result, to our best knowledge, C$^{3}$LR sets a new SoTA for most challenging settings of mini-ImageNet and CDFSL benchmarks.
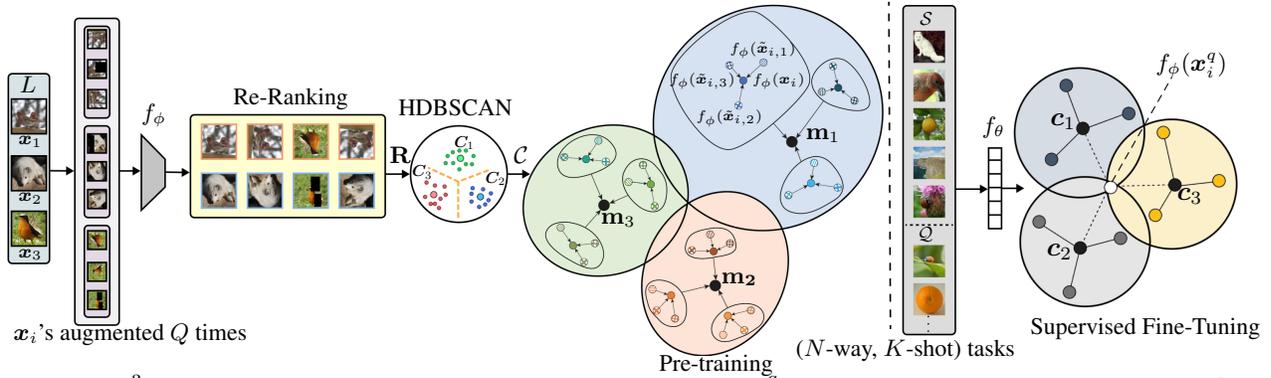
\begin{figure}
    \centering
    \begin{overpic}[abs, unit=1cm, width=\linewidth, trim=1.0cm 10.4cm 1cm 2.6cm, clip, percent]{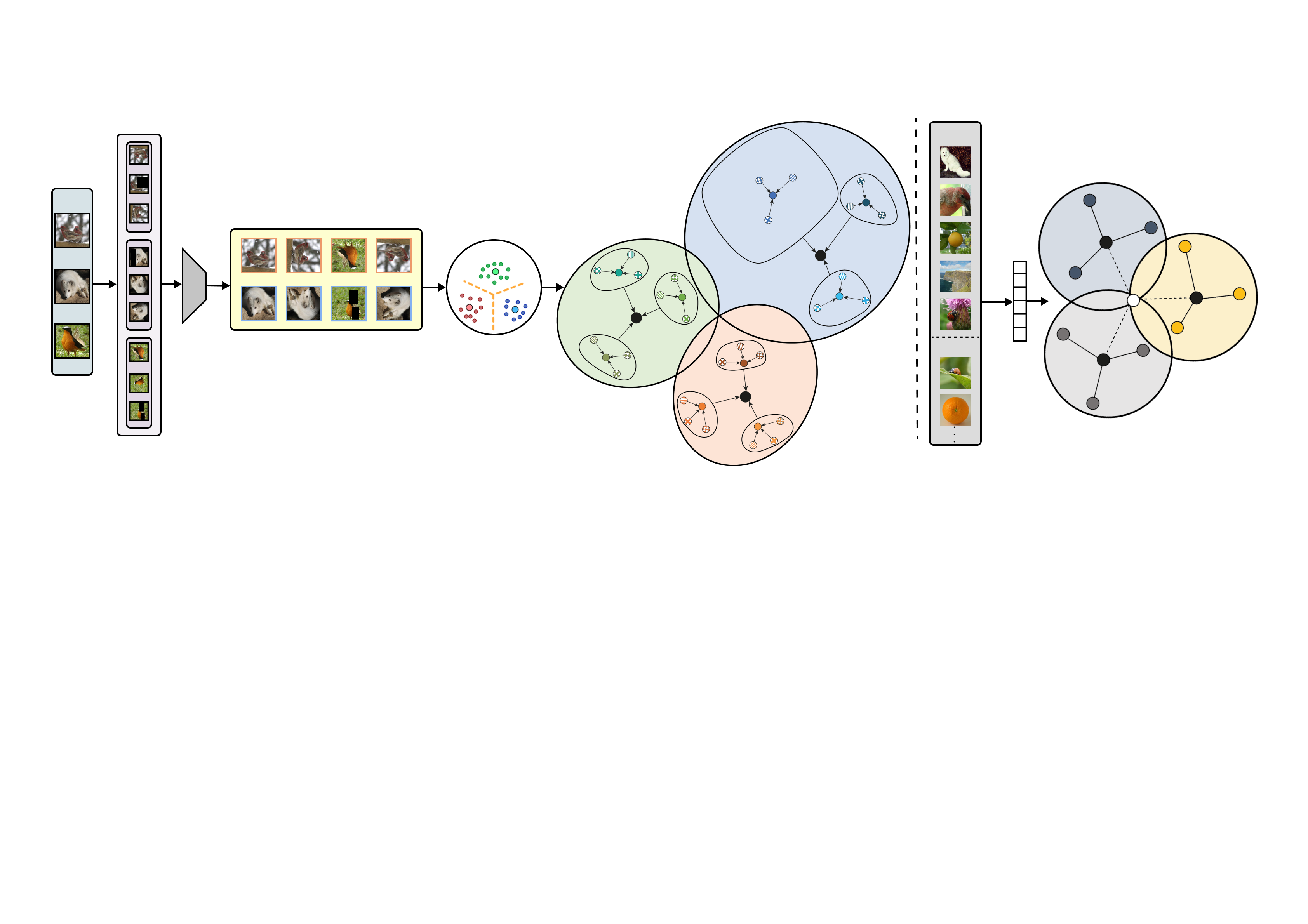}
    \put(1.5, 21.3){\footnotesize ${L}$}
    \put(1.4, 17.1){\scriptsize $\bm{x}_1$}
    \put(1.4, 12.5){\scriptsize $\bm{x}_2$}
    \put(1.4, 8){\scriptsize $\bm{x}_3$}

    \put(1, 1.3) {\footnotesize {$\bm{x}_i$'s augmented $Q$ times}}
    
    \put(11.3, 19) {\footnotesize $f_\phi$}
    
    \put(18.7, 20.3) {\footnotesize Re-Ranking}
    \put(31.3, 15.6) {\footnotesize $\mathbf{R}$}
    
    \put(32, 19.4) {\footnotesize HDBSCAN}
    \put(36.5, 16.9) {\scalebox{.6}{$\bm{C}_1$}}
    \put(39, 14) {\scalebox{.6}{$\bm{C}_2$}}
    \put(33.1, 14.5) {\scalebox{.6}{$\bm{C}_3$}}
    \put(41.3, 15.6) {\footnotesize $\mathcal{C}$}
    \put(53, -1.2) {\footnotesize Pre-training}
    \put(64.8, 17.5) {\small $\mathbf m_1$}
    \put(60.5, 22) {\scalebox{0.9}{\tiny $f_\phi(\bm{x}_i)$}}
    \put(58.7, 24.5) {\scalebox{0.9}{\tiny $f_\phi(\tilde{\bm{x}}_{i,1})$}}
    \put(56, 19) {\scalebox{0.9}{\tiny $f_\phi(\tilde{\bm{x}}_{i,2})$}}
    \put(53.8, 22) {\scalebox{.9}{\tiny $f_\phi(\tilde{\bm{x}}_{i,3})$}}
    \put(58, 5.9) {\small $\mathbf{m_2}$}
    \put(48.3, 10.7) {\small $\mathbf m_3$}
    
    \put(74, 26.7) {\scalebox{.7}{$\mathcal{S}$}}
    \put(73.8, 9.4) {\scalebox{0.7}{$\mathcal{Q}$}}
    \put(64, 0) {\footnotesize{($N$-way, $K$-shot) tasks}}
    \put(82.9, 1.8) {\footnotesize Supervised Fine-Tuning}

    \put(79, 18) {\footnotesize $f_\theta$}
    
    \put(84.5, 18.6) {$\bm c_1$}
    \put(84.2, 8.1) {$\bm c_2$}
    \put(94.7, 12.6) {$\bm c_3$}
    \put(89.4, 14){\tikz \draw[densely dashed,black] (0.8,0)--(1.6, 1.4);}
    \put(93, 23) {\small $f_\phi(\bm{x}_i^q)$}
    
    \end{overpic}
    \caption{C$^3$LR schematic view and training procedure. In the figure, $\bm{x}_i^q$ is an image sampled from the query set $\mathcal{Q}$. }
    
    \label{fig:flow}
\end{figure}

\section{Class-Cognizant Contrastive Learning (C$^3$LR)}\label{sec:c3lr-algo}
In this section, we first describe our problem formulation. We then discuss the two phases of the proposed approach: self-supervised pre-training and few-shot supervised fine-tuning. The mechanics of the proposed approach and a sketch of the training procedure is shown in \Cref{fig:flow}.   

\subsection{Preliminaries}\label{sse:preliminaries}

Let us denote the training data of size $M$ as $\mathcal{D}_{\textup{tr}} = \{({\bm x}_i, y_i)\}_{i = 1}^{M}$ with $({\bm x}_i, y_i)$ representing an image ${\bm x}_i$ and its class label $y_i$. In the pre-training phase, we take $L$ random samples from $\mathcal{D}_{\textup{tr}}$ and augment each sample $Q$ times by drawing augmentation functions $\psi^q(.), \forall q \in [Q]$ from the set $\mathcal{A}$. This results in a batch of size $B = (Q + 1)L$ total samples. Note that the data labels are unknown in the pre-training phase. In the fine-tuning phase, we deal with the so-called episodic training on a set of tasks $\mathcal{T}$ containing $N$ classes each with $K$ samples per task drawn from the test dataset $\mathcal{D}_{\textup{tst}} = \{({\bm x}_i, y_i)\}_{i = 1}^{M'}$ of size $M'$. From now on, we refer to this task construct as ($N$-way, $K$-shot) denoted by $(N, K)$. An episode consists of a labeled support set, $\mathcal{S}$, from which the model learns and an unlabeled query set, $\mathcal{Q}$, on which the model predicts. Note that both $\mathcal{S}$ and $\mathcal{Q}$ contain a set of tasks of the form $(N, K)$.

\subsection{Self-Supervised Pre-Training}
\label{ssec:pretraining}
The fact that we do not have access to class labels calls for a self-supervised pre-training stage. As discussed earlier, we build upon the idea of employing contrastive learning for prototypical transfer learning following the footsteps of \citep{Medina2020Self-SupervisedClassification}. The \textbf{high-level idea} here is to not only enforce the latent embeddings of augmented images come close to that of the source image in the embedding space (the classical contrastive setting), but also enforce embeddings of the images belonging to each cluster (and their augmentations) come closer to each other, for which a preceding unsupervised cluster formation step is required. This can help enforce similar classes into separate clusters, which will in turn be used as additional information in a modified two-term contrastive loss in \Cref{alg:pclr-c}. Let us walk you through the process in further details.

\begin{algorithm}[t!]
    \caption{\scalebox{.9}{Class-Cognizant Contrastive Learning (C$^3$LR)}}\label{alg:pclr-c}
    
    \SetKwInOut{Input}{input}
    \SetKwInOut{Output}{output}
    \SetKwInput{Require}{Require}
	\SetKwInput{Return}{Return}
	\SetKw{Let}{let}
	\SetKwRepeat{Do}{do}{while}
	
	\SetAlgoLined
	\LinesNumbered
	\DontPrintSemicolon
	\SetNoFillComment
    \Require{$L$, $Q$, $f_\phi$, $\mathcal{A}$, $\alpha$, $d[\cdot, \cdot]$}
    
    \While {not done}{
        Sample minibatch $\left\{\boldsymbol{x}_{i}\right\}_{i=1}^{L}$\;
        \ForAll{$i \in\{1, \ldots, L\}$} {
            \ForAll{$q \in\{1, \ldots, Q\}$}{
                $\tilde{\boldsymbol{x}}_{i, q}=\psi^{q}(\boldsymbol{x}_{i})$; $\psi^{q} \sim \mathcal{A}$.\;  
            }
        }
        $\mathbf{R} = \texttt{ReRank}$\scalebox{.8}{$\left(\left[ f_{\phi }\left(\{\boldsymbol{x}_{i}\}_{i=1}^{L}\right) ,f_{\phi }\left(\left\{\tilde{\boldsymbol{x}}_{i,\ q}\right\}_{i=1,q=1}^{L,Q}\right)\right]\right)$}\;

        $\mathcal{C} = \{\bm{C}_1 ,\bm{C}_2 ,\dotsc ,\bm{C}_{P}\} \gets \texttt{HDBSCAN}(\mathbf{R})$\;

        $\mathcal{M} = \{{\bf m}_{p}\}_{p = 1}^P;$ \hspace{0.1cm} ${\bf m}_p = \frac{\sum _{x_{j} \in \bm{C}_{p}} x_{j}}{|\bm{C}_{p} |}$\;
        
        \vspace{0.15cm}
        \Let{\scalebox{.95}{$\mathscr{r} (i,q,p)=-\log\frac{\exp\left( -d\left[ f_\phi\left(\tilde{{\boldsymbol x}}_{i,q}\right) ,\bf{m}_{p}\right]\right) }{\sum_{p=1}^{P}\exp\left( -d\left[ f_\phi\left(\tilde{\boldsymbol{x}}_{i,q}\right) ,\bf{m}_{p}\right]\right)}$}}\;
        \vspace{+0.15cm}
        \Let{\scalebox{.95}{$\ell(i, q)=-\log \frac{\exp \left(-d\left[f_\phi\left(\tilde{{\boldsymbol x}}_{i, q}\right), f_\phi\left(\boldsymbol{x}_{i}\right)\right]\right)}{\sum_{k=1}^{L} \exp \left(-d\left[f_\phi\left(\tilde{{\boldsymbol x}}_{i, q}\right), f_\phi\left(\boldsymbol{x}_{k}\right)\right]\right)}$}}\;
        
        \vspace{+0.15cm}
        $\mathcal{L}_{1}=\frac{1}{L Q} \sum_{p=1}^{P} \sum_{i=1}^{L} \sum_{q=1}^{Q} \mathscr{r}(i, q, p)$\;
        $\mathcal{L}_{2}=\frac{1}{L Q} \sum_{i=1}^{L} \sum_{q=1}^{Q} \ell(i, q)$\;
        
        $\mathcal{L} = \mathcal{L}_{1} + \mathcal{L}_{2}$
        
        $\phi \gets \phi-\alpha \nabla_{\phi} \mathcal{L}$\;
    }
\end{algorithm}

\Cref{alg:pclr-c} starts with \textbf{batch generation} (lines $2$ to $7$): each mini-batch consists of $L$ random samples $\left\{\boldsymbol{x}_{i}\right\}_{i=1}^{L}$ from $\mathcal{D}_{\textup{tr}}$, where $\bm{x}_i$ is treated as a $1$-shot support sample for which we create $Q$ randomly augmented versions $\tilde{\bm{x}}_{i, q}$ as query samples (line $5$). This leads to a batch size of $B = (Q + 1)L$. 
Then embeddings are generated by passing the samples through an encoder $f_{\phi}$ network. This is where the first major modification to ProtoTransfer \citep{Medina2020Self-SupervisedClassification} comes into play. Before the contrastive loss comes into action, we apply \textbf{re-ranking and clustering} (lines $8$ to $10$) to discover class-level global structure of data and enforce similar classes into separate clusters in the embedding space. Note that this step remains to be unsupervised in that the class labels are not required. The re-ranking step (line $8$) makes use of the $k$-reciprocal nearest neighbors as the distance metric between latent embeddings \citep{ZhongRe-rankingEncoding}, which has been shown to outperform the Euclidean distance \citep{Ji2019UnsupervisedTraining} when used for subsequent clustering.  
HDBSCAN clustering \citep{McInnes2017Hdbscan:Clustering} is then applied on the re-ranked embeddings ${\bf R}$ and returns a set of clusters populated in $\mathcal{C}$. HDBSCAN is versatile enough to discover and create required number of clusters $P$. With clusters at hand, we are now in a position to extend the standard loss proposed in \citep{Medina2020Self-SupervisedClassification} to contain a \textbf{class-cognizant term} (in lines $11$ and $13$), with lines $12$ and $14$ reflecting on the classical contrastive loss of ProtoTransfer \citep{Medina2020Self-SupervisedClassification}. This new loss term $\mathcal{L}_1$ enables a progressive improvement in class-level cluster formation and in turn learning similar representations for cluster members, while $\mathcal{L}_2$ encourages clustering of the embeddings of the augmented query samples $\left\{f_{\phi}(\tilde{\bm{x}}_{i, q})\right\}$ around their prototypes $\left\{f_{\phi}(\bm{x}_i)\right\}$. Here, both terms use an Euclidean distance metric in the embedding space denoted by $d[\cdot, \cdot]$. Finally, the new loss $\mathcal{L} = \mathcal{L}_1 + \mathcal{L}_2$ is optimized with mini-batch stochastic gradient decent with respect to the parameters $\phi$ of the encoder networks $f_{\phi}$.

\subsection{Supervised Fine-Tuning}
\label{ssec:fine-tuning}
The pre-trained encoder $f_\phi$ will be used for the downstream few-shot classification task. To this aim, following \citep{Medina2020Self-SupervisedClassification, Snell2017PrototypicalLearning}, we concatenate $f_{\phi}$ with a single-layer nearest-neighbor classifier $f_{\theta}$ (resulting in a similar architecture as in ProtoNet \citep{Snell2017PrototypicalLearning}) and fine-tune this last layer. In this phase, we first calculate the class prototypes $\bm{c}_{n}$ (embeddings) for class $n$ using the encoder $f_\phi$ on the support set $\mathcal{S}_n$:
\begin{equation}
    \nonumber
    \abovedisplayskip=5pt
    \boldsymbol{c}_{n}=\frac{1}{|\mathcal{S}_{n}|} \sum_{({\bm x}_i, y_i) \in {\mathcal{S}_n}} f_{\phi}(\bm{x}_{i}).
    \belowdisplayskip=5pt
\end{equation}
These prototypes are then used to initialize the classifier $f_{\theta}$ following \citep{Medina2020Self-SupervisedClassification}. 

\section{Experimentation}\label{sec:experiments}
In this section, we first discuss our experimental setup; we then present our numerical results. 

\subsection{Experimental Setup}\label{ssec:exp-setup}
\textbf{Datasets.} We conduct several in-domain experiments to benchmark C$^3$LR. For this purpose, we make use of commonly adopted datasets Omniglot \citep{Lake2015Human-levelInduction} and mini-Imagenet \citep{Vinyals2016MatchingLearning} to compare against unsupervised few-shot learning approaches. 
Omniglot contains $1623$ different handwritten characters borrowed from $50$ unique alphabets out of which we use $1028$ characters for training, $172$ for validation and $423$ for testing. We  resize the grayscale images to $28 \times 28$ pixels. Mini-ImageNet contains $100$ classes with $600$ samples in each class amounting to a total of $60,000$ images that we resize to $84 \times 84$ pixels. Out of the $100$ classes, we use $64$ classes for training, $16$ for validation and  $20$ for testing. For both datasets, the settings are the most commonly adopted ones in literature \citep{Vinyals2016MatchingLearning, Medina2020Self-SupervisedClassification, Ji2019UnsupervisedTraining, Khodadadeh2018UnsupervisedClassification}. The augmentations (in $\mathcal{A}$) used for the experimentations follow \citep{Medina2020Self-SupervisedClassification}.
We also compare our method on a more challenging cross-domain few-shot learning (CDFSL) benchmark \citep{guo2019new}. This benchmark consists of four datasets with increasing similarities to mini-ImageNet. In that order, we have grayscale chest X-ray images from ChestX \citep{wang2017chestx}, dermatological skin lesion images from ISIC2018 \citep{isic2018dataset},  satellite aerial images from EuroSAT \citep{helber2017eurosat}, and crop disease images from CropDiseases \citep{mohanty2016using}. We also use Caltech-UCSD Birds (CUB) dataset \citep{WahCUB_200_2011} for further analysis of cross-domain performance. CUB is composed of $11,788$ images from 200 unique bird species. We use $100$ images for training, $50$ for validation and $50$ for test. 

\textbf{Training.} The Conv$4$ model \citep{Vinyals2016MatchingLearning} is pre-trained on the respective training splits of the datasets, with an initial learning rate of $0.001$, multiplied by $0.5$ every $25,000$ steps via the Adam optimizer \citep{kingma2014adam}. Based on the derivations in \citep{Snell2017PrototypicalLearning} and similar usage in \citep{Medina2020Self-SupervisedClassification}, we initialize the classification layer $f_{\theta}$ with weights set to $\mathbf{W}_{n}=2 \bm{c}_{n}$ and biases set to $b_{n}=-\left\|\bm{c}_{n}\right\|^{2}$. For validation, we create $15$ ($N$-way, $K$-shot) tasks using the validation split from which the corresponding validation accuracy and loss are calculated. Experiments involving CDFSL benchmark follow \citep{guo2019new, Medina2020Self-SupervisedClassification}, where we pre-train a ResNet$10$ encoder using C$^3$LR on mini-ImageNet images of size $224 \times 224$ for $400$ epochs with the Adam optimizer and a constant learning rate of $0.001$.

\textbf{Evaluation scenarios and baseline.} Our testing scheme uses $600$ test episodes on which the pre-trained encoder (using \ccclr) is fine-tuned and tested. All our results indicate $95\%$ confidence intervals over $3$ runs each with $600$ test episodes. The standard deviation values are thus calculated according to the $3$ runs to provide more solid measures for comparison. For our in-domain benchmarks, we test on ($5$-way, $1$-shot) and ($5$-way, $5$-shot) classification tasks. While our cross-domain testing is done using ($5$-way, $5$-shot) and ($5$-way, $20$-shot) classification tasks. We compare our performance with a suit of recent self-supervised few-shot baselines such as ProtoTransfer \citep{Medina2020Self-SupervisedClassification}, UFLST \citep{Ji2019UnsupervisedTraining}, LASIUM \citep{Khodadadeh2020UnsupervisedModels} and CACTUS \citep{Hsu2018UnsupervisedMeta-Learning}, to name a few. Furthermore, we also compare with a set of supervised approaches (such as MAML \citep{Finn2017Model-agnosticNetworks}, ProtoNet \citep{Snell2017PrototypicalLearning} , etc.) the best performing of which are obviously expected to outperform ours as well as other self-supervised methodologies.
\begin{table}[t!]
	\small
    \caption{Accuracy ($\% \pm$ std.) for ($N$-way, $K$-shot) classification tasks. Style: \textbf{best} and \underline{second best}.}
    \centering
	{\tabcolsep=0pt\def\arraystretch{1.1}
		\begin{tabularx}{\linewidth}{l *2{@{}>{\centering\arraybackslash}X} | *2{@{}>{\centering\arraybackslash}X}}
        \toprule
        &\multicolumn{2}{c |}{\bf Omniglot} & \multicolumn{2}{c}{\bf mini-ImageNet}
        \tabularnewline \cmidrule(lr){2-3}\cmidrule(l){4-5}
        {\bf Method$(N,K)$} & {(5,1)} & {(5,5)}  & {(5,1)} & {(5,5)} \\ 
        
        \midrule
        
        CACTUs-MAML \citep{Hsu2018UnsupervisedMeta-Learning}     & 68.84 \scriptsize{$\pm$ 0.80}         & 87.78 \scriptsize{$\pm$ 0.50}                                   & 39.90 \tiny{$\pm$ 0.74}    & 53.97 \scriptsize{$\pm$ 0.70} \\
        CACTUs-ProtoNet \citep{Hsu2018UnsupervisedMeta-Learning} & 68.12 \scriptsize{$\pm$ 0.84}         & 83.58 \scriptsize{$\pm$ 0.61}                                   & 39.18 \scriptsize{$\pm$ 0.71}    & 53.36 \scriptsize{$\pm$ 0.70}\\
        UMTRA \citep{Khodadadeh2018UnsupervisedClassification}   & 83.80           & 95.43                                      & 39.93 & 50.73 \\
        AAL-ProtoNet \citep{Antoniou2019AssumeAugmentation}      & 84.66 \scriptsize{$\pm$ 0.70}         & 89.14 \scriptsize{$\pm$ 0.27}                                   & 37.67 \scriptsize{$\pm$ 0.39} & 40.29 \scriptsize{$\pm$ 0.68} \\
        AAL-MAML++ \citep{Antoniou2019AssumeAugmentation}        & 88.40 \scriptsize{$\pm$ 0.75}         & \underline{97.96} \scriptsize{$\pm$ 0.32}                       & 34.57 \scriptsize{$\pm$ 0.74} & 49.18\scriptsize{$\pm$ 0.47} \\
        UFLST \citep{Ji2019UnsupervisedTraining}                 & \textbf{97.03}   & \textbf{99.19}                             & 33.77 \scriptsize{$\pm$ 0.70} & 45.03 \scriptsize{$\pm$ 0.73}\\
        ULDA-ProtoNet \citep{Qin2020DiversityAugmentation} & - & - & 40.63 \scriptsize{$\pm$ 0.61} & 55.41 \scriptsize{$\pm$ 0.57}\\
        ULDA-MetaOptNet \citep{Qin2020DiversityAugmentation} & - & - & 40.71 \scriptsize{$\pm$ 0.62} & 54.49 \scriptsize{$\pm$ 0.58} \\
        U-SoSN+ ArL \citep{Zhang2020RethinkingLearning} & - & - & 41.13 \scriptsize{$\pm$ 0.84} & 55.39 \scriptsize{$\pm$ 0.79} \\
        LASIUM \citep{Khodadadeh2020UnsupervisedModels} & 83.26 \scriptsize{$\pm$ 0.55} & 95.29 \scriptsize{$\pm$ 0.22} & 40.19 \scriptsize{$\pm$ 0.58} & 54.56 \scriptsize{$\pm$ 0.55} \\
        ProtoTransfer {\scriptsize{($L=50$)}}  \citep{Medina2020Self-SupervisedClassification}  & 88.00 \scriptsize{$\pm$ 0.64}         & 96.48 \scriptsize{$\pm$ 0.26}               & \underline{45.67} \scriptsize{$\pm$ 0.79}        & \underline{62.99} \scriptsize{$\pm$ 0.75}\\
        ProtoTransfer \scriptsize{($L=200$)}                                        & 88.37 \scriptsize{$\pm$ 0.74}         & 96.54 \scriptsize{$\pm$ 0.41}               & 44.17 \scriptsize{$\pm$ 1.08} & 61.07 \scriptsize{$\pm$ 0.82} \\
    	\rowcolor{teal!20}C$^3$LR (\textbf{ours}) & \underline{89.30} \scriptsize{$\pm$ 0.64} & 97.38 \scriptsize{$\pm$ 0.23} & \textbf{47.92} \scriptsize{$\pm$ 1.2}  & \textbf{64.81} \scriptsize{$\pm$ 1.15} \\
        \cdashlinelr{1-5}
        {MAML \citep{Finn2017Model-agnosticNetworks} (supervised)} & 94.46 \scriptsize{$\pm$ 0.35} & 98.83 \scriptsize{$\pm$ 0.12} & 46.81 \scriptsize{$\pm$ 0.77} & 62.13\scriptsize{$\pm$ 0.72} \\
        {ProtoNet \citep{Snell2017PrototypicalLearning} (supervised)} & 97.70\scriptsize{$\pm$ 0.29} & 99.28 \scriptsize{$\pm$ 0.10} & 46.44\scriptsize{$\pm$ 0.78}  & 66.33\scriptsize{$\pm$ 0.68} \\
        {MMC \citep{ren2018meta}  (supervised)} & 97.68\scriptsize{$\pm$ 0.07} & - & 50.41 \scriptsize{$\pm$ 0.31} & 64.39 \scriptsize{$\pm$ 0.24}\\
        {FEAT \citep{ye2020few} (supervised)   } & - & - & 55.15 & 71.61 \\
        {Pre+Linear \citep{Medina2020Self-SupervisedClassification}  (supervised)} & 94.30 \scriptsize{$\pm$ 0.43} & 99.08 \scriptsize{$\pm$ 0.10} & 43.87 \scriptsize{$\pm$ 0.69} & 63.01 \scriptsize{$\pm$ 0.71} \\
        \bottomrule
        \end{tabularx}}
    \label{tab:ResultsOmniglotMini}
\end{table}
\begin{table}[t!]
    \scriptsize
    \caption{Accuracy ($\% \pm$ std.) for ($N$-way, $K$-shot) classification on mini-ImageNet with pre-training on CUB.}

    {\tabcolsep=0pt\def\arraystretch{1}
    \begin{tabularx}{\linewidth}{l *3{@{}>{\centering\arraybackslash}X}}
        \toprule
        {\bf Training} & {\bf Testing} & {(5,1)} & {(5,5)} \\
        \midrule
        ProtoTransfer \scriptsize{($L=50$)} \citep{Medina2020Self-SupervisedClassification} & ProtoTune \citep{Medina2020Self-SupervisedClassification} & \underline{35.37} \scriptsize{$\pm$ 0.63} & \underline{52.38} \scriptsize{$\pm$ 0.66} \\
        ProtoTransfer \scriptsize{($L=200$)}  & ProtoTune & 34.67 \scriptsize{$\pm$ 0.84} & 51.45 \scriptsize{$\pm$ 0.72} \\
        \rowcolor{teal!20}C$^3$LR (\textbf{ours}) & ProtoTune & \textbf{39.61} \scriptsize{$\pm$ 1.11} & \textbf{55.53} \scriptsize{$\pm$ 1.42} \\
        \bottomrule
        
        \end{tabularx}}
    \label{tab:results_cub2mini}
\end{table}
\begin{table}[th!]
    \scriptsize
	\caption{Accuracy ($\% \pm$ std.) of ($N$-way, $K$-shot) classification on the CDFSL benchmark. Style: \textbf{best} and \underline{second best}.}
            {\tabcolsep=0pt\def\arraystretch{1.0}
            \begin{tabularx}{\linewidth}{@{}>{\arraybackslash}l *2{@{}>{\centering\arraybackslash}X} | *2{>{\centering\arraybackslash}X} | *2{>{\centering\arraybackslash}X} | *2{>{\centering\arraybackslash}X}}
				\toprule
				{\bf Method$(N,K)$}            & {(5,5)}                 & {(5,20)}                                & {(5,5)}        & {(5,20)}        & {(5,5)}        & {(5,20)}  & {(5,5)}        & {(5,20)}                  \\ 
				\midrule
				            
				& \multicolumn{2}{c}{\bf ChestX} & \multicolumn{2}{|c}{\bf{ISIC}} & \multicolumn{2}{|c}{\bf EuroSAT}  & \multicolumn{2}{|c}{\bf CropDiseases} \\
				            
				\midrule
				UMTRA-ProtoNet \citep{Medina2020Self-SupervisedClassification}              & 24.94 \scriptsize{$\pm$ 0.43}             & 28.04 \scriptsize{$\pm$ 0.44}              & 39.21 \scriptsize{$\pm$ 0.53}             & 44.62 \scriptsize{$\pm$ 0.49}     & 74.91 \scriptsize{$\pm$ 0.72}             & 80.42 \scriptsize{$\pm$ 0.66}              & 79.81 \scriptsize{$\pm$ 0.65}             & 86.84 \scriptsize{$\pm$ 0.50}              \\
				UMTRA-ProtoTune \citep{Medina2020Self-SupervisedClassification}                & 25.00 \scriptsize{$\pm$ 0.43}             & 30.41 \scriptsize{$\pm$ 0.44}              & 38.47 \scriptsize{$\pm$ 0.55}             & 51.60 \scriptsize{$\pm$ 0.54}  & 68.11 \scriptsize{$\pm$ 0.70}             & 81.56 \scriptsize{$\pm$ 0.54}              & 82.67 \scriptsize{$\pm$ 0.60}             & 92.04 \scriptsize{$\pm$ 0.43}              \\
				ProtoTransfer \citep{Medina2020Self-SupervisedClassification}                   & \textbf{26.71} \scriptsize{$\pm$ 0.46}    & \textbf{33.82} \scriptsize{$\pm$ 0.48}     & \underline{45.19} \scriptsize{$\pm$ 0.56}             & \underline{59.07} \scriptsize{$\pm$ 0.55}      & \underline{75.62} \scriptsize{$\pm$ 0.67}             & \underline{86.80} \scriptsize{$\pm$ 0.42}              & \underline{86.53} \scriptsize{$\pm$ 0.56}             & \underline{95.06} \scriptsize{$\pm$ 0.32}  \\
				\rowcolor{teal!20}C$^3$LR (\textbf{ours})                   &  \underline{26.00} \scriptsize{$\pm$ 0.41}            & \underline{33.39} \scriptsize{$\pm$ 0.47}  & \textbf{45.93} \scriptsize{$\pm$ 0.54}            &  \textbf{59.95} \scriptsize{$\pm$ 0.53}     & \textbf{80.32} \scriptsize{$\pm$ 0.65}             & \textbf{88.09} \scriptsize{$\pm$ 0.45}     &   \textbf{87.90} \scriptsize{$\pm$ 0.55}             & \textbf{95.38} \scriptsize{$\pm$ 0.31}              \\
				\cdashlinelr{1-9}
				ProtoNet \citep{guo2019new} (sup.)                      & 24.05 \scriptsize{$\pm$ 1.01}             & 28.21 \scriptsize{$\pm$ 1.15}              & 39.57 \scriptsize{$\pm$ 0.57}             & 49.50 \scriptsize{$\pm$ 0.55}       & 73.29 \scriptsize{$\pm$ 0.71}             & 82.27 \scriptsize{$\pm$ 0.57}              & 79.72 \scriptsize{$\pm$ 0.67}             & 88.15 \scriptsize{$\pm$ 0.51}              \\
				Pre+Mean-Cent. \citep{guo2019new} (sup.)             & 26.31 \scriptsize{$\pm$ 0.42} & 30.41 \scriptsize{$\pm$ 0.46}             & 47.16 \scriptsize{$\pm$ 0.54} & 56.40 \scriptsize{$\pm$ 0.53}              & 82.21 \scriptsize{$\pm$ 0.49}    & 87.62 \scriptsize{$\pm$ 0.34}              & 87.61 \scriptsize{$\pm$ 0.47}             & 93.87 \scriptsize{$\pm$ 0.68}              \\
				Pre+Linear \citep{guo2019new} (sup.)                  & 25.97 \scriptsize{$\pm$ 0.41}             & 31.32 \scriptsize{$\pm$ 0.45}              & 48.11 \scriptsize{$\pm$ 0.64}    & 59.31 \scriptsize{$\pm$ 0.48}   & 79.08 \scriptsize{$\pm$ 0.61}             & 87.64 \scriptsize{$\pm$ 0.47}  & 89.25 \scriptsize{$\pm$ 0.51}    & 95.51 \scriptsize{$\pm$ 0.31}     \\

				\bottomrule
			\end{tabularx}}
	\label{tab:cdfsl_full}
\end{table}

\subsection{Performance Evaluation}\label{ssec:perf-eval}

\textbf{In-domain evaluation.} \Cref{tab:ResultsOmniglotMini} summarizes our performance evaluation results on Omniglot and mini-ImageNet datasets for ($N$-way, $K$-shot) scenarios with $N = 5$ and $K = 1, 5$. The top section compares the performance of the proposed approach (C$^3$LR) with the most recent relevant self-supervised competitors. As can be seen, for Omniglot, we outperform ProtoTransfer \citep{Medina2020Self-SupervisedClassification} (which we build on) by about $1\%$ in both $K=1, 5$ shot scenarios. We score the second overall best in ($5$-way, $1$-shot) falling behind UFLST \citep{Ji2019UnsupervisedTraining}. For the mini-ImageNet benchmark, to our knowledge, we set a new SoTA outperforming ProtoTransfer by $2\%+$. Interestingly, our performance beats some of the supervised baselines (bottom section of the table) adopting similar encoder architecture Conv$4$ for mini-ImageNet and comes close to $K = 5$-shot performances on Omniglot. Obviously, the SoTA supervised few-shot learning approaches have the advantage of having access to the all the labels, as such due to the supervision signal, are expected to outperform the unsupervised approaches like ours.    

\textbf{Cross-domain evaluation.} So far we have demonstrated that the proposed approach excels for in-domain scenarios. The next step is to assess the performance under more challenging cross-domain scenarios (\Cref{tab:results_cub2mini} and \Cref{tab:cdfsl_full}) where we pre-train on a certain dataset in an unsupervised fashion, then fine-tune and test on a different dataset. \Cref{tab:results_cub2mini} illustrates the results of a Conv$4$ encoder trained on CUB and tested on tasks derived from mini-ImageNet. Here again C$^3$LR shows a clear improvement of $3\%+$ compared to ProtoTransfer (with pre-training sample sizes $L = 50, 200$). The important message here is that the proposed approach enhances ProtoTransfer in generalizing to truly unseen data. To further investigate the performance on cross-domain scenarios, we next focus on CDFSL benchmark \citep{guo2019new} containing several datasets. Here, we pre-train on mini-ImageNet and fine-tune and test on ChestX \citep{wang2017chestx}, ISIC2018 \citep{isic2018dataset}, EuroSAT \citep{helber2017eurosat}, and CropDiseases \citep{mohanty2016using}. We compare the performance against ProtoTransfer and two of its variants with UMTRA \citep{Khodadadeh2018UnsupervisedClassification} as pre-training strategy (all proposed in \citep{Medina2020Self-SupervisedClassification}). We also compare with a couple of closely related supervised approaches from \citep{guo2019new}, for the sake of reference. As can be seen, except for ChestX where we marginally come short of ProtoTransfer, for the other three datasets we outperform the second best competitor (ProtoTransfer) by about $0.5\%+$ to $4.5\%+$ with the most significant improvement in the case of EuroSAT. Interestingly, once again the performance of C$^3$LR is not far off that of the related supervised approaches (bottom of the table) even sometimes outperforming the supervised approaches especially in ($5$-way, $20$-shot) scenarios.

\section{Concluding Remarks}\label{sec:conclusion}

Inspired by the idea of using contrastive learning for unsupervised few-shot classification, we build upon the recently proposed idea of ProtoTransfer \citep{Medina2020Self-SupervisedClassification} by incorporating class cognizance through: (i) an unsupervised iterative re-ranking and clustering step, followed by (ii) an adjusted optimization loss formulation. We demonstrate that our proposed approach (C$^3$LR) offers considerable performance improvement above its predecessor ProtoTransfer in both in/cross-domain few-shot classification scenarios setting a new SoTA in mini-ImageNet and CDFSL benchmarks.

\bibliographystyle{unsrtnat}
\bibliography{references}  

\end{document}